\documentclass[10pt, conference, letterpaper]{IEEEtran}
\IEEEoverridecommandlockouts

\usepackage{balance}
\usepackage{amssymb}
\usepackage{stmaryrd}
\usepackage{cite}
\usepackage{color}
\usepackage{multirow}
\usepackage{graphicx,times}
\usepackage{epstopdf}
\usepackage{indentfirst}
\usepackage{CJK}
\usepackage{amsmath}
\usepackage{amsfonts}
\usepackage{txfonts}
\usepackage{mathrsfs}
\usepackage{theorem}
\usepackage{float}
\usepackage{subfigure}

\usepackage{verbatim}
\usepackage{hhline}
\usepackage{algorithmic}
\usepackage{algorithm}
\usepackage{array}
\usepackage{url}
\usepackage{bm}
\usepackage{textcomp}
\usepackage{xcolor}
\usepackage{microtype}

\def\BibTeX{{\rm B\kern-.05em{\sc i\kern-.025em b}\kern-.08em
    T\kern-.1667em\lower.7ex\hbox{E}\kern-.125emX}}

\newcommand{\ie}{{\em i.e.}}
\newcommand{\eg}{{\em e.g.}}

\def\spth{\textsuperscript{th}}

\DeclareMathOperator*{\argmin}{argmin}

\begin{document}
\title{Localizing Backscatters by a Single Robot With Zero Start-up Cost}

\author{\IEEEauthorblockN{Shengkai Zhang{$^{^\dagger}$}, Wei Wang{${^\ast}^{^\dagger}$}, Sheyang Tang{${^{^\dagger}}$}, Shi Jin{${^\S}$}, Tao Jiang{$^{^\dagger}$}}\IEEEauthorblockA{{$^{^\dagger}$}School of Electronic Information and Communications, Huazhong University of Science and Technology \\{${^\S}$} School of Information Science and Engineering, Southeast University\\Email: \{szhangk, weiwangw, sheyangtang, taojiang\}@hust.edu.cn, jinshi@seu.edu.cn}
\thanks{${^\ast}$The corresponding author is Wei Wang (weiwangw@hust.edu.cn). }
}

\maketitle

\begin{abstract}
Recent years have witnessed the rapid proliferation of low-power backscatter technologies that realize the ubiquitous and long-term connectivity to empower smart cities and smart homes. Localizing such low-power backscatter tags is crucial for IoT-based smart services. However, current backscatter localization systems require prior knowledge of the site, either a map or landmarks with known positions, increasing the deployment cost. To empower universal localization service, this paper presents Rover, an indoor localization system that simultaneously localizes multiple backscatter tags with zero start-up cost using a robot equipped with inertial sensors. Rover runs in a joint optimization framework, fusing WiFi-based positioning measurements with inertial measurements to simultaneously estimate the locations of both the robot and the connected tags. Our design addresses practical issues such as the interference among multiple tags and the real-time processing for solving the SLAM problem. We prototype Rover using off-the-shelf WiFi chips and customized backscatter tags. Our experiments show that Rover achieves localization accuracies of $39.3$ cm for the robot and $74.6$ cm for the tags. 
\end{abstract}

\begin{IEEEkeywords}
	Backscatter, localization, inertial sensor, channel state information
\end{IEEEkeywords}

\section{Introduction}
\label{sec:intro}
The last few years have seen rapid innovations in low-power backscatter communication~\cite{kellogg2014wi, peng2018plora, elmossallamy2018backscatter}. These designs enable wireless devices to communicate at microwatts of power and operate reliably at moderate ranges to provide whole-home or warehouse coverage. A more recent design~\cite{hessar2019netscatter} enables low-power backscatter networks to support hundreds to thousands of concurrent transmissions. Such devices will be a key enabler of ubiquitous sensing and interconnection in the era of Internet-of-Things (IoT) due to its low cost, small footprint and battery-free communications~\cite{li2018game}. 

Localizing low-power backscatters is crucial for many IoT applications, such as battery-free network maintenance~\cite{hessar2019netscatter} and universal object localization~\cite{kotaru2017localizing}. Unfortunately, past work of backscatter localization has one of two shortcomings: 1) It either requires the deployment of multiple landmarks~\cite{kotaru2017localizing}, which usually costs much effort to pre-calibrate the positions of these landmarks to set up a coordinate system, or, 2) it needs additional sensing modalities, such as visual sensing~\cite{ma2017drone, lin2018autonomous}, to enable the RF localizability.

Ideally, we desire a system that supports the IoT localization and extends the tracking demand from smartphones and wearables to universal objects, such as wallets, keys, and pill bottles: the system should be power-on-and-go that works without any effort for start-up, \eg, site survey or landmark position setup, and it should be ubiquitously available that works with low-power sensor suite upon existing infrastructure for rapid deployment and long-term service. 

\begin{figure}[t!]
  \centering
  \includegraphics[width=2.4in]{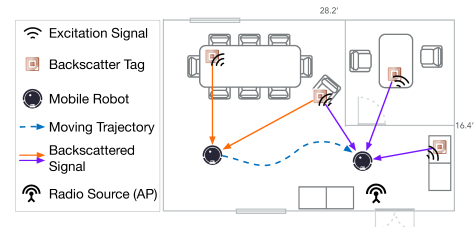}
  \setlength{\belowcaptionskip}{-16pt}
  \caption{A robot moves in a conference room to simultaneously localize itself and the backscatters who connect to it during the movement. For large-scale backscatter networks, the backscatters out of the robot's communication range can be localized when the robot moves nearby.}
  \label{fig:toy}
\end{figure}

In this paper, we propose Rover, a power-on-and-go low-power backscatter localization system that works with existing ubiquitous infrastructure, \ie, commodity WiFi. Rover supports instant deployment with zero start-up cost. It is a self-contained system that runs on a robot equipped with WiFi chips and an inertial measurement unit (IMU). Rover simultaneously localizes the robot and the backscatter tags that communicate with the robot. It needs to rove in the work space to connect to more tags for localization as shown in Fig.~\ref{fig:toy}. The core of Rover is a simultaneous-localization-and-mapping (SLAM) approach that leverages spatially different observations to construct multi-view constraints for localizing the map points and the observer, corresponding to the tags and the robot in our system.

Despite the advantages of Rover, there are two significant challenges. 
{\em First}, without knowing the positions of a robot or any tag, enough translations and angle-of-arrivals (AoAs) of a tag to the robot at different positions are required to satisfy the requirement of triangulation. A straightforward method to obtain the translation is integrating the accelerations measured by IMU. However, the integration operation will lead to a temporal drift due to the inherent sensor noise~\cite{he2017pervasive}. Fortunately, the AoA we obtained is drift-free for localization. We use it to correct the IMU drift by proposing an AoA-IMU SLAM system that jointly optimizes the locations of the robot and the connected backscatter tags with WiFi AoAs and IMU measurements. {\em Second}, the optimization framework takes the measurements from WiFi and IMU at different locations and our goal is to find a configuration of such locations that best fit all these measurement constraints. In principle, taking more measurements over the robot's trajectory into account for the optimization would provide more accurate results. However, this also incurs more complex computations and delays the localization, making the robot unable to navigate itself when moving. To bound the computation complexity for real-time processing while achieving better accuracies, Rover employs a sliding window based formulation for the SLAM problem and derives the solution in a graph-based optimization framework. 

{\bf Results}. We prototype Rover on the programmable robot, iRobot Create 2, equipped with an IMU and an Intel Next Unit of Computing (NUC) that installs an Intel 5300 wireless NIC. We use the 802.11n Channel State Information (CSI) tool~\cite{halperin2011tool} to obtain wireless channel information for AoA estimation. To implement the backscatter tag, we use the hardware provided by HitchHike~\cite{zhang2016hitchhike} and reprogram its FPGA with our Rover firmware. The experiments are conducted with four backscatters deployed in a conference room to validate individual system modules as well as the overall performance. The results show that Rover achieves localization accuracies of $74.6$ cm for the backscatter tag and $39.3$ cm for the robot over a trajectory of $41.96$ m.

{\bf Contributions}. Rover is the first low-power backscatter localization system that works with a single robot using commodity WiFi without any prior knowledge of work space. Rover leverages the localizability of WiFi signals to correct the IMU drift. {\em First}, we propose an AoA-IMU SLAM system that jointly optimizes the locations of the robot and the connected backscatter tags. {\em Second}, we employ a sliding window based formulation for the SLAM problem to achieve real-time and accurate processing. We implement Rover on commodity devices and experimentally validate the system in indoor environments.

\section{Backscatter AoA Estimation}
\label{sec:background}
\begin{figure}[t!]
    \centering
    \shortstack{
            \includegraphics[width=0.23\textwidth]{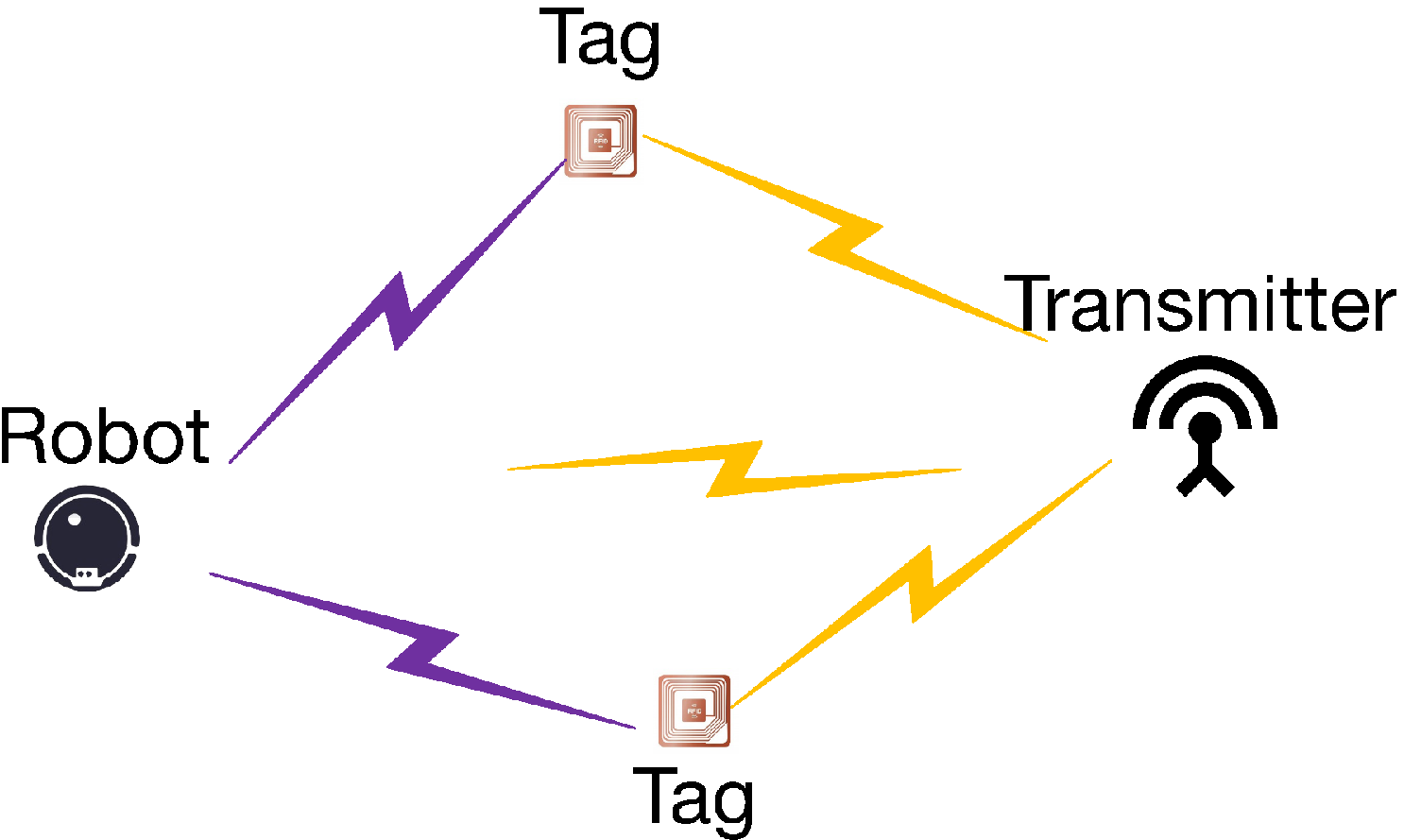}\\
            {\footnotesize (a) Signal interference}
    }
    \shortstack{
            \includegraphics[width=0.23\textwidth]{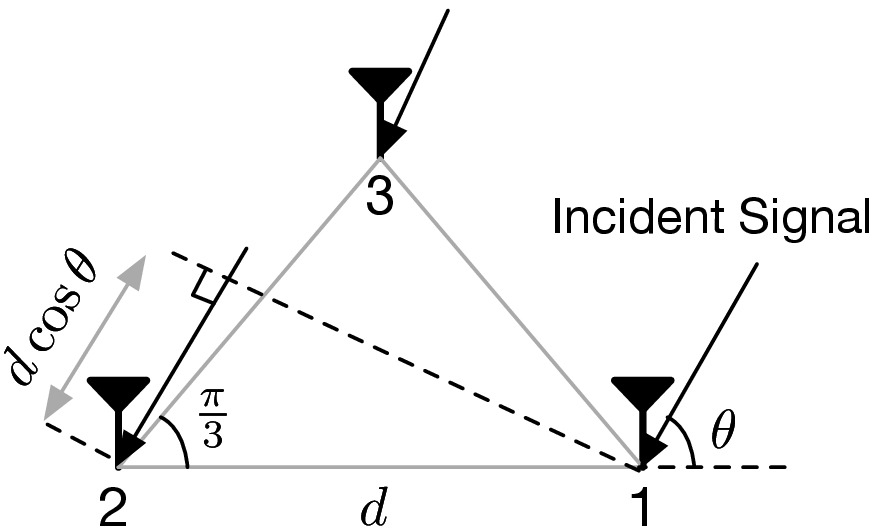}\\
            {\footnotesize (b) Uniform circular array}
    }
    \caption{(a) shows the interference caused by the backscattered signals (purple flash) from tags and the excitation signal (yellow flash) from a transmitter. (b) shows a uniform circular array consisting of three antennas: The signal with AoA $\theta$ travels an additional distance of $d\cos\left(\theta + \frac{\pi}{3}\right)$ to the third antenna and $d\cos(\theta)$ to the second array in the array compared to the first antenna.}
    \label{fig:background}
\end{figure}

Differing from conventional WiFi localization systems that the target device responds a WiFi receiver through an active WiFi radio, backscatter localization systems arise two problems: 1) The WiFi receiver should be able to decode the backscattered low-power signals from tags; 2) Multiple tags that concurrently backscatter signals can interfere with each other or with the WiFi transmitter as shown in Fig.~\ref{fig:background}~(a). The key to address the first problem is to make sure the tag reflects the preamble of the excitation WiFi packet. We implement an envelop detector to detect the starting point of a WiFi packet and backscatter a decodable WiFi packet as proposed in~\cite{kotaru2017localizing}. To resolve the interference problem, frequency shifting~\cite{zhang2016hitchhike} is effective that we can build a tag that shifts the WiFi signal by a particular frequency to another channel and then backscatters the signal. Multiple tags need to shift into different channels. While the number of WiFi channels is limited, we propose an interference avoidance mechanism that allows Rover to work in a scalable battery-free network.  

\subsection{Interference Avoidance}
To avoid the interference from a WiFi transmitter, we toggle the RF transistor of a backscatter at a higher speed~\cite{zhang2016hitchhike}, \eg, $20$ MHz. Then, the backscatter signal will be moved to a channel that is $20$ MHz away from the channel where the excitation WiFi signal stays. Configuring the receiver to work on the channel of the backscatter signal will address the interference from the excitation signal.

To avoid the interference from multiple tags that concurrently backscatter signals, the above idea can be extended to assign separate WiFi channels to each of the tags by toggling their RF transistors to different speeds. Then, the receiver needs to sweep all WiFi channels except the channel of the excitation signal to localize multiple tags. We achieve this by implementing a frequency band sweeping protocol~\cite{vasisht2016decimeter} in the iwlwifi driver of Intel 5300 NIC. Since the number of non-overlapped WiFi bands is limited, Rover can only simultaneously localize a limited number of tags. To maximize the ability of simultaneous localization, it is vital to choose the channel of the excitation signal. 

Suppose a tag uses a frequency $f_b$ square wave signal to control the on-off frequency of the RF switch. $f_c$ is the carrier center frequency of the 802.11n excitation signal. Let $\omega_b = 2\pi f_b$, $\omega_c = 2\pi f_c$, and $\alpha_{\text{base}}(t)$ denotes a baseband waveform. The square wave can be formulated as $S_{\text{tag}}(t) = \frac{4}{\pi}\sum_{n=1}^{\infty}\frac{\sin\left[(2n-1)\omega_b t\right]}{2n-1}$. Hence, the backscatter signal $\beta(t)$ can be written as,
\begin{equation}
	\beta(t) = \alpha_{\text{base}}(t)e^{j\omega_c t}S_{\text{tag}}(t).
\end{equation}
Let $F_{\text{base}}(\omega)$ and $F(\omega)$ be the Fourier transform of $\alpha_{\text{base}}(t)$ and $\beta(t)$ respectively. We have 
\begin{equation}
	\begin{aligned}
		F(\omega) =  & \sum_{n=1}^{\infty}\frac{2j}{\pi (2n-1)}\left( F_{\text{base}} \left( \omega - \omega_c + (2n-1)\omega_b \right) - \right. \\
		& \left. F_{\text{base}} \left( \omega - \omega_c - (2n-1)\omega_b \right) \right).
	\end{aligned}
\end{equation}
This indicates that the frequency-shifted backscattered signal can be received at two bands, $f_c \pm f_b$, causing sideband interference to other channels. Based on this, Rover sets the most side channels in the band, \ie, channel $165$ or $36$, to avoid the sideband interference. 

\subsection{AoA Estimation}
So far, we have addressed the interference problem. A WiFi receiver can receive the backscatter packets from tags. In this section, we describe how Rover estimates the AoAs of backscatter tags to the receiver, which is amounted on a robot, leveraging the CSI of received packets. The AoA estimation technique for low-power backscatter tags was first proposed in~\cite{kotaru2017localizing}. Here we extend it to work with a circular antenna array with uniform spacing $d$ that can measure AoAs in $[0, 360]$ degrees as shown in Fig.~\ref{fig:background}~(b).

\begin{figure}[t!]
  \centering
  \includegraphics[width=2.4in]{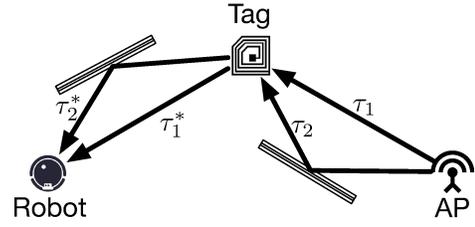}
  \setlength{\belowcaptionskip}{-16pt}
  \caption{The received signal traverses two physical paths where $\tau_j$ denotes the time of flight (ToF) of $j$\spth path on the transmitter-to-tag link and $\tau^*_i$ denotes the traversing time of $i$\spth path on the tag-to-receiver link.}
  \label{fig:paths}
\end{figure}

Localizing backscatter tags involves two physical paths, transmitter-to-tag path and tag-to-receiver path, as shown in Fig.~\ref{fig:paths}. Thus, the received CSI depends on the locations of backscattered tags and the access point (AP). We combine $j$\spth path on the transmitter-to-tag link with $i$\spth path on the tag-to-receiver link to form a virtual path between the AP and the receiver at the robot. The virtual path has a time-of-flight (ToF) of $\widehat{\tau}_k = \tau_j + \tau_i^*$ where $\tau_i^*$ ($\tau_j$) denotes the ToF of the signal along $i$\spth($j$\spth) path on the tag-to-receiver (transmitter-to-tag) link, the AoA of the virtual path $\widehat{\theta}_k = \theta_i^*$ where $\theta_i^*$ is the AoA of $i$\spth path on the tag-to-receiver link, and the corresponding complex attenuation of $\widehat{\gamma}_k = \gamma_j \gamma_i^*$ where $\gamma_i^*$ and $\gamma_j$ denote the complex attenuation along $i$\spth path on the tag-to-receiver link and $j$\spth path on the transmitter-to-tag link, respectively. The overall signal obtained at the three antennas for $n$\spth subcarrier can be written as
\begin{equation}
	\begin{aligned}
		H_{n, m} & = \sum_{k=1}^{L_{\text{tx}} L_{\text{tag}} }\widehat{\gamma}_k e^{-j2\pi\left(\widehat{\tau}_k(n-1)f_\delta + (m-1)d \cos\widehat{\theta}_k/\lambda\right)}, m = 1, 2	\\
		H_{n, 3} & = \sum_{k=1}^{L_{\text{tx}} L_{\text{tag}} }\widehat{\gamma}_k e^{-j2\pi\left(\widehat{\tau}_k(n-1)f_\delta + d \cos\left(\widehat{\theta}_k+\frac{\pi}{3}\right)/\lambda\right)},
	\end{aligned}
\end{equation}
where $L_{\text{tag}}$ is the number of paths on the tag-to-receiver link, $L_{\text{tx}}$ the number of paths on the transmitter-to-tag link, $f_\delta$ the frequency gap between two consecutive subcarriers. This overall signal is reported as CSI corresponding to the particular subcarrier and antenna.

The signal model is a standard form to apply the joint AoA-ToF estimation technique~\cite{kotaru2015spotfi}. The insight of this technique is that multiple subcarriers of an OFDM signal encode ToF information. By smoothing the subcarriers represented in the CSI matrix, it allows a super-resolution AoA estimation with a small antenna array, \eg, a three-antenna array available for Intel 5300 NIC, jointly estimating AoAs and ToFs\footnote{This ToF cannot correctly infer the traveling distance of a propagation path due to its poor distance resolution from the narrowband signal.} of all paths. The AoA of the path with the smallest ToF is the direct-path AoA of a tag to the receiver. 

At this stage, we obtain the direct-path AoAs of multiple tags to the receiver. Next, we fuse them with the IMU measurements to localize the tags and the robot simultaneously.

\section{SLAM with AoAs}
\label{sec:design}

Without knowing the position of any device, how can we localize the target by a single mobile robot? In this section, we answer this question by first describing the design of our localization system and then elaborating on the sliding window based solution.

\subsection{AoA-IMU Localization System}
The angle (AoA) can be used to determine a target's location via triangulation. Recall that conventional localization systems usually require a few landmarks with known locations to localize the target via angles. The essence of this requirement is defining the metric scale of environments, \ie, the unit (meter, millimeter, etc.) in measuring distances between objects, to fix the size of triangles. 

\begin{figure}[t!]
  \centering
  \includegraphics[width=2.4in]{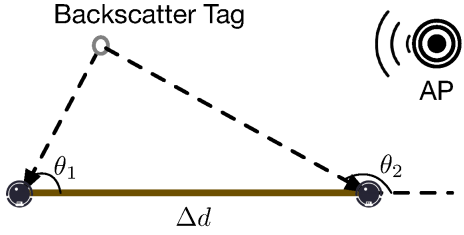}
  \setlength{\belowcaptionskip}{-16pt}
  \caption{Localization principle: triangulation with the robot's motions. The AP sends WiFi packets to excite the backscatter tag. The receiver on the robot measures the AoAs from backscatter signals and the onboard IMU measures translation $\Delta d$ to provide the metric scale of environments.}
  \label{fig:principle}
\end{figure}

In Rover, since the location of both the tags and the robot are unknown, the AoA we obtained cannot yield locations with the metric scale of environments. Nevertheless, with the aid of the onboard IMU and the mobility of a robot, we can localize both the connected tags and the robot in that the IMU provides accelerations in unit $m/s^2$, defining the metric scale. In addition, with motions and AoAs of the incident signals at different positions, it meets the triangulation principle. We take one tag as an example illustrated in Fig.~\ref{fig:principle}. As a robot moves, the IMU measures translation $\Delta d$ and the antenna array measures AoAs $\theta_1$ and $\theta_2$ referring to the tag at different positions, one can determine the relative positions of the robot and the tag through triangulation. 

Obtaining the translation by integrating the accelerations from the IMU is straightforward but suffers from temporal accumulated errors due to the inherent noise~\cite{he2017pervasive}, causing large localization errors once the result severely distorts the triangle in Fig.~\ref{fig:principle}. To address this issue, we develop an AoA-IMU SLAM approach that optimizes the locations of the robot and backscatters subject to measurement constraints with respect to WiFi AoAs and the IMU odometry. Roughly speaking, the central idea of SLAM is to obtain a maximum likelihood estimate of both robot positions and environment features (backscatter tags in our system) given observations (AoAs from the antenna array). Solutions to the SLAM problem can be either filtering-based or graph-based approaches. While filtering-based approaches are considered to be more efficient in computation, we choose graph-based approaches that can achieve better performance via repetitively linearizing past robot states and multi-view constraints~\cite{lin2018autonomous}. 

In addition, solving the SLAM problem is a batch process that incorporates multiple observations to produce accurate results. However, it can become unacceptably slow as the size of the environment grows. This delays the location estimates of the robot so that the robot loses its own navigation capability, being unable to move along the desired trajectory. To let our system run in real-time, we employ an incremental update method to speed up the computation. We formulate a sliding window based model that only keeps a limited amount of AoAs and corresponding robot hidden {\em states}, \eg, the positions of the robot at different timestamps in the workspace, to bound the computation complexity. 

\subsection{Sliding Window Formulation}
\begin{figure}[t!]
  \centering
  \includegraphics[width=2.5in]{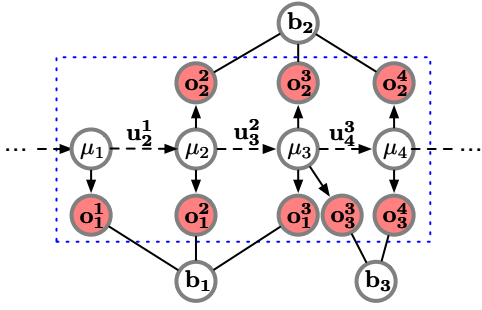}
  \caption{Graph representation for the sliding window SLAM. $\bm{\mu}$ is the hidden state of robot position; $\mathbf{b}$ denotes the hidden state of backscatter position; $\mathbf{o}$ denotes the AoA observation; $\mathbf{u}$ is the odometry captured by IMU. The sliding window represented by the blue dashed box contains four states and their observed AoAs.}
  \label{fig:estimator}
\end{figure}

Fig.~\ref{fig:estimator} shows the graph representation of our SLAM formulation. Let $\bm{\mu}_i$ denote the hidden state at discrete timestamp $i$. At each timestamp, the robot observes a set of AoAs from multiple backscatters. $\mathbf{o}_{i}^j$ is the AoA of backscatter $j$ observed at timestamp $i$ and $\mathbf{b}_j$  denotes the position of backscatter $j$. The relative translation between two robot states $\bm{\mu}_i$ and $\bm{\mu}_{i+1}$ is captured by an odometry edge $\mathbf{u}_{i+1}^i$, which can be obtained by IMU preintegration techniques~\cite{forster2015rss}. 

We define a state vector in the sliding window that merges the hidden variables of robot and backscatter together, 
\begin{equation}   
  		\bm{\mathcal{S}} 	= [\bm{\mu}_0, \bm{\mu}_1, \dotsc, \bm{\mu}_{n-1}, \mathbf{b}_0, \mathbf{b}_1, \dotsc, \mathbf{b}_{m-1}]^\top, 
\label{eqn:state}
\end{equation}
\noindent where the initial position $\bm{\mu}_0 = [0, 0, 0]$. All these variables refer to the world frame, which maps to the real world where the gravity is vertical. $n$ is the number of robot's state in the sliding window, $m$ the number of observed backscatter tags, and $\mathbf{b}_i$ the position of tag $i$ in the world frame. At this stage, we have constructed the graph from the AoA observations and the IMU odometry measurements. Next step, we seek to find the configuration of the positions of the robot and tags that best satisfies the constraints, \ie, the edges in the graph.

Since our system only involves translations, parameters in $\bm{\mathcal{S}}$ are in Euclidean space. We can formulate the problem as a linear system and the optimal state sequence $\bm{\mathcal{S}}^*$ in the sliding window can be estimated by solving:
\begin{equation}
		\bm{\mathcal{S}}^* = \argmin_{\bm{\mathcal{S}}} \Big\{\overbrace{\mathbf{A}(\bm{\mathcal{S}})}^\text{AoA constrains} + \overbrace{\mathbf{D}(\bm{\mathcal{S}})}^\text{odometry constrains} \Big\},
  	\label{eqn:cost}
\end{equation}
where 
\begin{equation}
	\begin{aligned}
		\mathbf{A}(\bm{\mathcal{S}}) &=  \sum_{(i, j)\in\mathcal{A}}\left\|\hat{\mathbf{o}}_{i}^{j} - \mathbf{Q}_{i}^{j}\bm{\mathcal{S}}\right\|^2_{\bm{\Omega}_{i}^j} \\
		\mathbf{D}(\bm{\mathcal{S}}) & = \sum_{k\in\mathcal{I}}\left\|\hat{\mathbf{u}}_{k+1}^{k} - \mathbf{P}_{k+1}^{k}\bm{\mathcal{S}}\right\|^2_{\bm{\Lambda}_{k+1}^k}.
	\end{aligned}
  	\label{eqn:cost_specific}
\end{equation}
\noindent $\mathcal{A}$ denotes the set of AoA measurements between all tags and the robot in the window. $\mathcal{I}$ denotes the set of all inertial measurements in the window. To solve this system, the terms of the AoA constraints $\left\{\hat{\mathbf{o}}_{i}^{j}, \mathbf{Q}_{i}^{j}, \mathbf{\Omega}_{i}^j\right\}$ and the odometry constraints $\left\{\hat{\mathbf{u}}_{k+1}^{k}, \mathbf{P}_{k+1}^{k}, \bm{\Lambda}_{k+1}^k\right\}$ need to be defined.

{\bf AoA constraints}. The direction vector $\mathbf{r}_{i}^{j}$ referred to the observed $i$\spth tag at timestamp $j$ can be defined by the AoA $\theta$\footnote{We consider 2D cases for ease of presentation. Our system can be trivially extended to work in 3D space by using a larger antenna array or other sensors, \eg, ultrasonic sensor. Since the 3D extension is incremental to our contribution, we omit the details in this paper.} as $\mathbf{r}_{i}^{j} = [\cos(\theta), \sin(\theta), 0]^{\top}$. With an unknown distance $d_i^j$, a simple geometric relationship can be expressed as
\begin{equation}
  d_i^{j}\mathbf{r}_{i}^{j} = \mathbf{R}_{0}^{j}\left(\mathbf{b}_{i} - \bm{\mu}_{j}\right),
  \label{eqn:similar_equation}
\end{equation}
\noindent where $\mathbf{b}_{i}$ is the $i$\spth tag's position and $\bm{\mu}_j$ the robot position at timestamp $j$. Since $\mathbf{r}_{i}^{j}$ should have the same direction as the vector $\mathbf{b}_{i} - \bm{\mu}_{j}$ if the measurement noise is absent. The expected observation can be expressed by a cross product operation,
\begin{equation} 
  \hat{\mathbf{o}}_i^{j} = \hat{\mathbf{0}} = \left(\mathbf{R}_{j}^{0}\mathbf{r}_{i}^{j}\right) \times \left(\mathbf{b}_{i} - \bm{\mu}_{j}\right) = \mathbf{Q}_i^{j}\bm{\mathcal{S}} + \mathbf{n}_i^{j},
  \label{eqn:wifi_measurement_model}
\end{equation}
\noindent where $\mathbf{n}_i^{j}$ denotes the noise, assuming that it follows a Gaussian distribution. The AoA covariance $\bm{\Omega}_i^j$ can be pre-measured by statistical methods and updated along the optimization process. Specifically, $\bm{\Omega}_i^j = {d_{i}^{j}}^2 \bar{\bm{\Omega}}_i^j$, $d_{i}^{j}$ is the distance from the robot to $i$\spth tag at timestamp $j$ and $\bar{\bm{\Omega}}_i^j$ denotes the AoA observation noise. Initially, the distance $d_{i}^{j}$ is given by a reasonable guess. Then it will be refined automatically along the sliding window optimization as the positions of the robot and tags are updated. Therefore, the initial guess is insensitive.

{\bf Odometry constraints}. Typically, the data rate of IMU is higher than AoA rate. Given two consecutive timestamps $[k, k+1]$ at which the AoAs from multiple tags are received, there have been multiple buffered inertial measurements, which include acceleration $\mathbf{a}_t \in \mathbb{R}^{3}$ and angular rate $\bm{\omega}_t \in \mathbb{R}^{3}$. We can preintegrate them to obtain an overall odometry representation between $\bm{\mu}_k$ and $\bm{\mu}_{k+1}$ as follows:
\begin{equation}
  \begin{aligned}
  	\mathbf{V}_{k+1}^{k}  &= \sum_{t\in[k, k+1]}\mathbf{R}_t^{k}\mathbf{a}_t \Delta t     \\
    \mathbf{T}_{k+1}^{k} &= \sum_{t\in[k, k+1]}\left[\mathbf{V}_{k+1}^{k}\Delta t + \mathbf{R}_t^{k}\mathbf{a}_t\Delta t^2\right],
  \end{aligned}
  \label{eqn:preimu}
\end{equation}
\noindent where $\mathbf{R}_t^{k} = \sum_{i \in [k, t]}\left[\mathbf{R}_i^{k}\lfloor\bm{\omega}_t\times\rfloor \Delta t\right]$,  $\mathbf{R}_t^{k} \in \text{SO}(3)$. $\lfloor\bm{\omega}_t\times\rfloor$ is the skew-symmetric matrix from $\bm{\omega}_t$, $\Delta t$ the time interval between two consecutive measurements. $\mathbf{R}_t^{k}$ denotes the incremental rotation from time $k$ to current time $t$, which is available through short-term integration of gyroscope measurements. Then, we can write the propagation model of positions as
\begin{equation}
    \bm{\mu}_{k+1}  = \bm{\mu}_{k} + \mathbf{R}_{k}^0\bm{\nu}_{k}\Delta t - \mathbf{R}_{k}^{0}\textbf{g}\Delta t^2/2 + \mathbf{R}_{k}^0\mathbf{T}_{k+1}^{k}, 
  \label{eqn:linear_update}
\end{equation}
\noindent where $\textbf{g} = [0, 0, 9.8]^\top$ is the vertical gravity. Since the robot only moves in a room (assuming a horizontal plane), it is safe to obtain the accelerations that account for motions by subtracting the gravity. $\bm{\nu}_k$ denotes the velocity at timestamp $k$. Its propagation model can be expressed as 
\begin{equation}
	\bm{\nu}_{k+1} = \mathbf{R}_{k}^{k+1}\bm{\nu}_{k} - \mathbf{R}_{k}^{k+1}\textbf{g}\Delta t + \mathbf{R}_{k}^{k+1}\mathbf{V}_{k+1}^{k}.
	\label{eqn:velocity_update}
\end{equation}
$\mathbf{R}_{k}^{0}$ is the change in rotation since the initial state. We can see that the update equation for the quantity $\bm{\mu}_{k}$ and $\bm{\nu}_{k+1}$ will be linear in Eqn.~\eqref{eqn:linear_update} and Eqn.~\eqref{eqn:velocity_update} if rotation $\mathbf{R}_{k}^{0}$ are provided. This rotation can be obtained by solving a linear system that incorporate short-term integration of gyroscope measurements. For brevity, we omit the details and refer to the broad literature discussing these ideas~\cite{shen2016initialization}.

Accordingly, Eqn.~\eqref{eqn:linear_update} can be rewritten as a linear function of state $\bm{\mathcal{S}}$: 
\begin{equation}
  \begin{aligned}
    \hat{\mathbf{u}}_{k+1}^{k} = \hat{\mathbf{T}}_{k+1}^{k}   
    & = \mathbf{R}_{0}^{k}\left(\bm{\mu}_{k+1} - \bm{\mu}_{k}\right) - \bm{\nu}_{k}\Delta t + \textbf{g}\frac{\Delta t^2}{2}  \\
    &= \mathbf{P}_{k+1}^{k}\bm{\mathcal{S}} + \mathbf{n}_{k+1}^{k},
  \end{aligned}
  \label{eqn:imumodel}
\end{equation}
\noindent where $\bm{\nu}_k$ can be updated by Eqn.~\eqref{eqn:velocity_update}, $\mathbf{n}_{k+1}^{k}$ denotes the additive measurement noise. Typically, we assume the additive noise follows a Gaussian distribution. Then the covariance $\bm{\Lambda}_{k+1}^{k}$ can be calculated using the technique proposed in~\cite{lupton2012visual}.

At this point, all constraints in Eqn.~\eqref{eqn:cost_specific} are explicitly defined. The information matrices and state vectors in the sliding window can be stacked to construct a large array of linear equations so that the positions of the robot and tags in the window can be solved altogether. 

\section{Evaluation}
\label{sec:evaluation}
\subsection{Implementation and Experimental Setup}
\begin{figure}[t!]
  \centering
  \includegraphics[width=2.5in]{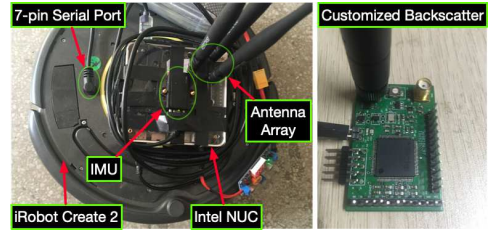}
  \caption{The experimental platform. The left shows the receiver attaches on the robot and sends commands to control its motions through the Create's 7-pin serial port. The right shows one of our customized backscatter tags.}
  \label{fig:sys}
\end{figure}

We implemented Rover on an Intel NUC with a 1.3 GHz Core i5 processor with $4$ cores, an $8$ GB of RAM and a $120$ GB SSD, running Ubuntu Linux equipped with Intel 5300 NICs and a LORD MicroStrain 3DM-GX4-45 IMU. We use the Linux 802.11 CSI tool~\cite{halperin2011tool} to obtain the wireless channel information for each packet. Thanks to the open-source hardware of HitchHike~\cite{zhang2016hitchhike}, we build the customized tags to backscatter commodity WiFi signals. The whole system is implemented in C++. The NUC connects to the iRobot Create 2 and uses ROS (Robot Operating System) as the interfacing robotics middleware to control the robot's moving trajectory\footnote{ROS driver for iRobot Create 2, \url{https://github.com/autonomylab/create_autonomy}.}. The experimental platform is shown in Fig.~\ref{fig:sys}.

In all experiments, we use two NUCs. One is the excitation source that operates in $5.825$ GHz center frequency (channel $165$) on a $20$ MHz band. The other is the receiver on the robot that performs the frequency hopping protocol to sweep all available channels in the $5$ GHz band except channel $165$. The experiments are conducted in a $9\times 5$ square meters meeting room in our laboratory, which is a typical indoor setting. Four backscatter tags are deployed in the room. Each tag is configured to shift a frequency and backscatter signals in a separate channel. This prevents the interference between tags. In addition, the frequency shift provides a mechanism to distinguish the received signal from which tag as each tag occupies a separate channel.

\subsection{AoA Estimation}
\begin{figure}[t!]
    \centering
    \shortstack{
            \includegraphics[width=0.16\textwidth]{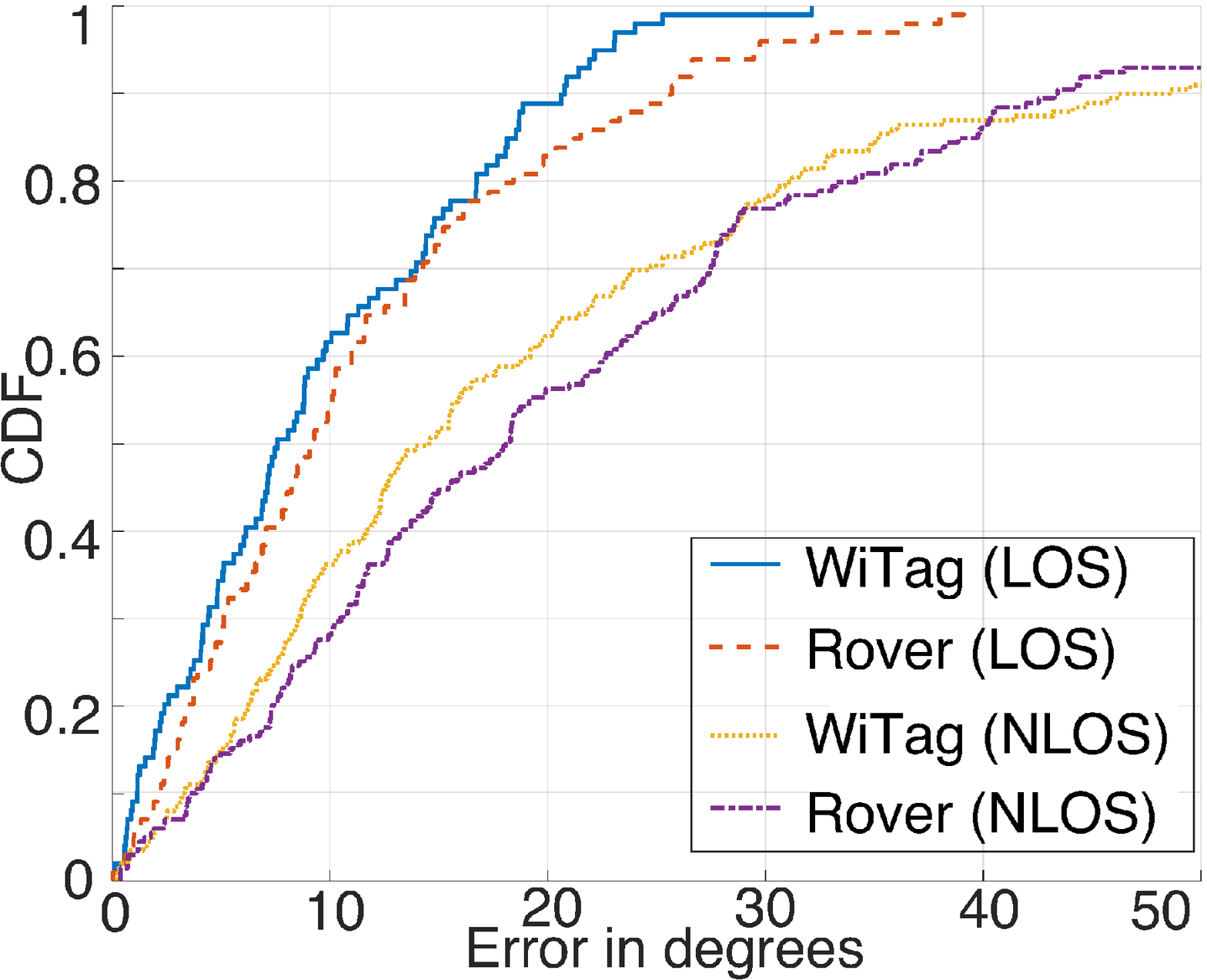}\\
            {\footnotesize (a) CDF of AoA error}
    }\quad
    \shortstack{
            \includegraphics[width=0.28\textwidth]{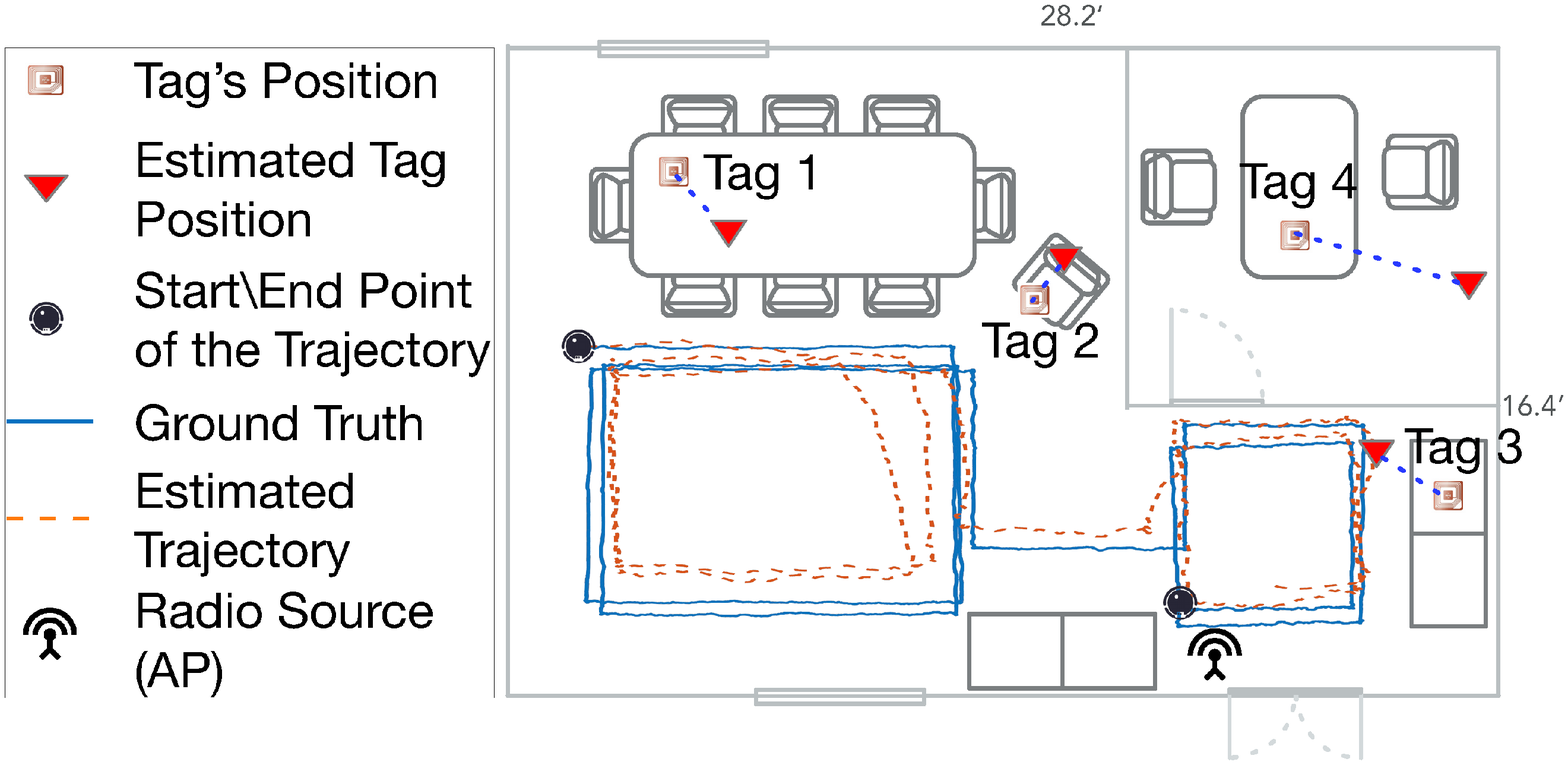}\\
            {\footnotesize (b) System deployment and estimated results}
    }
    \caption{(a) The accuracy of AoA estimation. (b) We use the NUC to control the robot moves in a pre-defined rectangular trajectory in the meeting room. The ground truth is provided by the program that defines the trajectory running in the NUC.}
    \label{fig:aoa} 
\end{figure}

The key difference in AoA estimation from the state-of-the-art~\cite{kotaru2017localizing}, WiTag, is that we empower it with time-division multiplexing so that the receiver can simultaneously measure AoAs of multiple tags who backscatter signals in different channels. We demonstrate the AoA estimation by four tags deployed in line-of-sight (LOS) and non-LOS (NLOS) settings. The CDF plotted in Fig.~\ref{fig:aoa} (a) shows that the performance of Rover is similar to WiTag. The median errors of Rover and WiTag are $9.3^{\circ}$ and $8.1^{\circ}$ respectively in LOS deployment. In NLOS deployment, the median errors of Rover and WiTag are $18.1^{\circ}$ and $14.6^{\circ}$, respectively.

\subsection{Simultaneous Localization}

\begin{figure}[t!]
    \centering
    \shortstack{
            \includegraphics[width=0.22\textwidth]{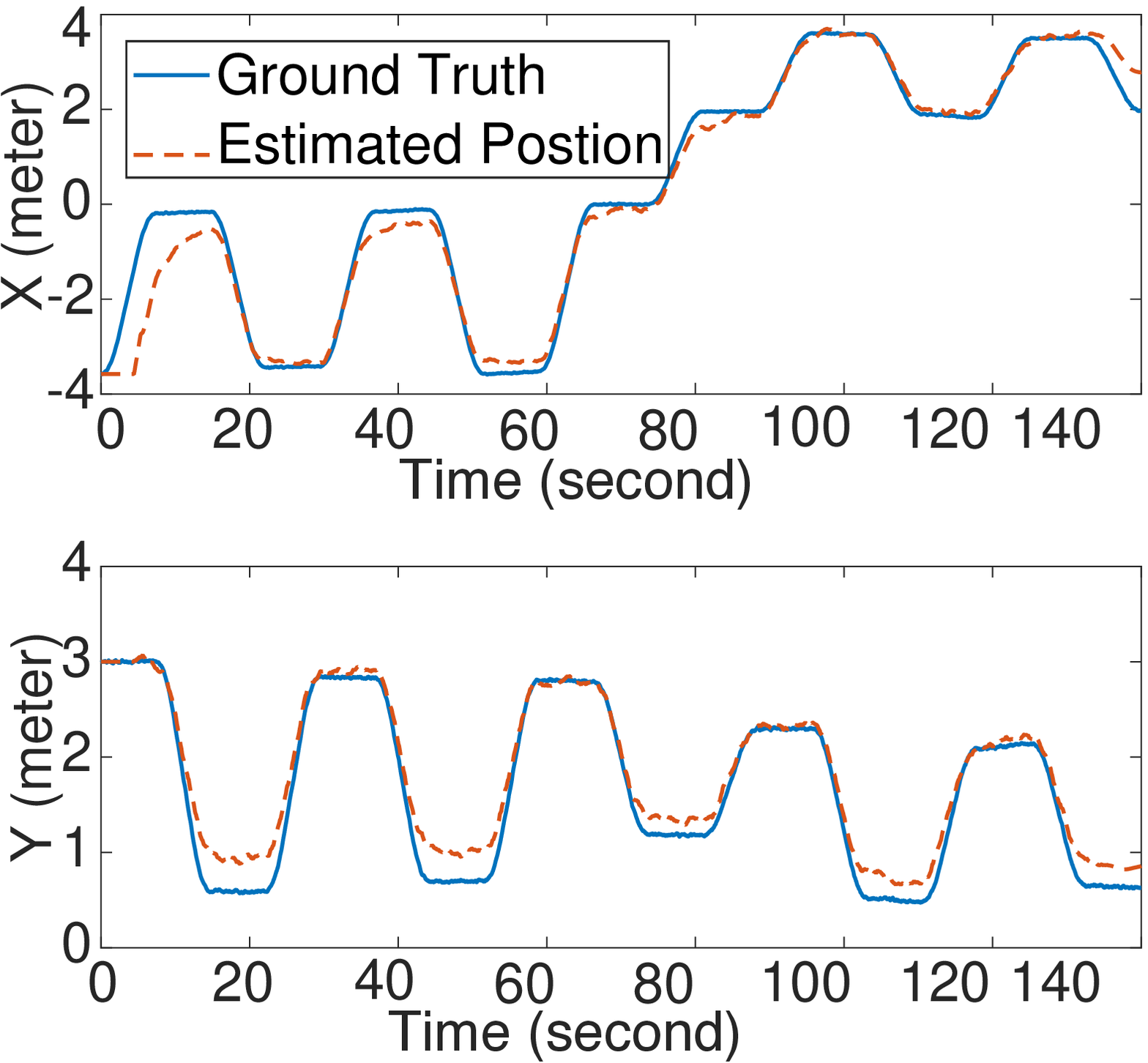}\\
            {\footnotesize (a) Robot position tracking}
    }\quad
    \shortstack{
            \includegraphics[width=0.22\textwidth]{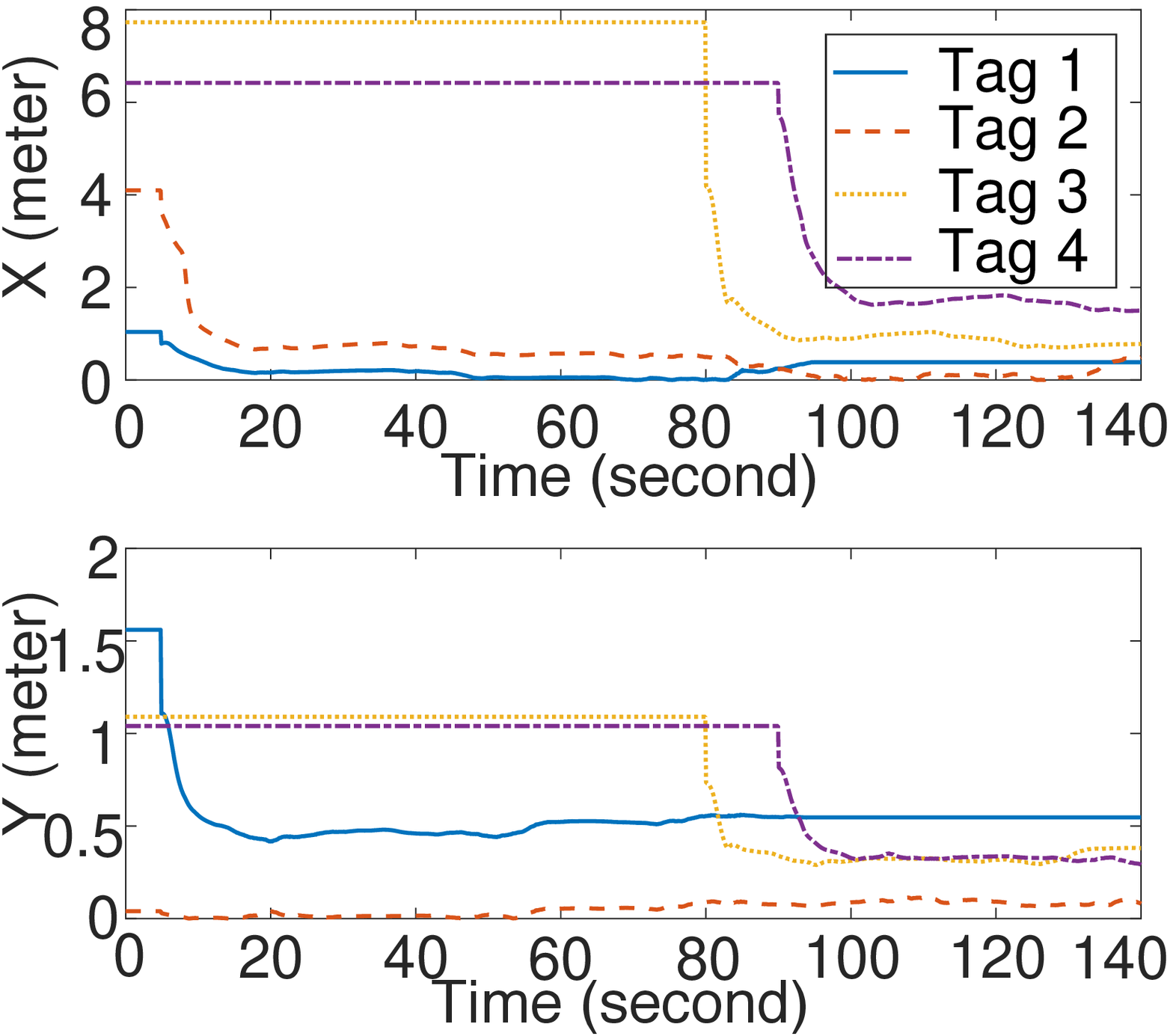}\\
            {\footnotesize (b) The localization error of four tags}
    }    
    \caption{SLAM performance in the meeting room.} 
    \label{fig:experiment_result}
\end{figure}

Fig.~\ref{fig:aoa}~(b) shows the system deployment and the overall performance of Rover. The performance of tracking the robot's trajectory is plotted in Fig.~\ref{fig:experiment_result} (a). The mean error over the estimated trajectory is $39.3$ cm. The accuracy goes beyond the expectation from the noisy AoA estimation (Fig.~\ref{fig:aoa}~(a)). This is because our system introduces the inertial sensors that provide an additional sensing modality for positioning. Moreover, our sliding window optimization filters out the noise of heterogeneous sensors by finding the configuration of positions that best fits different spatial measurement constraints. 

Fig.~\ref{fig:experiment_result} (b) shows the localization results of the tags. Initially, the tags' locations are set to be $(0, 0)$. After about $20$ seconds, Rover localizes tags $1$ and $2$ as their AoAs are available. Tags $3$ and $4$ are localized at about $80$ and $90$ seconds later as the robot approaches them and received their backscattered packets. Meanwhile, Rover stops updating the location of tag $1$ after about $95$ seconds as it loses contact with the robot. The four tags' final localization errors are $73.6$ cm, $52.9$ cm, $97.2$ cm, and $145.9$ cm, respectively. Among them, the error of tag $4$ is higher due to its NLOS deployment. The mean localization error in LOS deployment is $74.6$ cm.

In summary, the localization accuracy is decimeter-level, which is similar to the state-of-art WiFi based localization systems~\cite{kotaru2015spotfi, kotaru2017localizing}. The uniqueness of Rover is that it works without landmarks or any map of the environment, while conventional solutions need multiple APs with known positions. Conventional solutions use more APs to provide redundant positioning measurements and combat the noise of WiFi measurements. On the contrary, we take advantage of IMU and a robot's mobility to enable a new localization paradigm. The inertial measurements play the role of combating the WiFi noise and the drift-free WiFi measurements help correct the IMU drift in return.

\section{Conclusion}
\label{sec:conclusion}
We present Rover, a low-power backscatter localization system with an AoA-IMU SLAM framework. We formulate a sliding window based model that fuses inertial measurements with the AoAs of backscatter tags to a robot measured by commodity WiFi to simultaneously estimate the locations of the robot as well as the connected tags. We implement Rover on the iRobot Create 2 platform attached with an Intel NUC and an IMU. The experiments in both LOS and NLOS deployments in indoor settings show that Rover achieves localization accuracy of tens of centimeters for both the robot and the backscatter tags without any prior knowledge of the work space. Extending our system to work with other wireless devices, such as iBeacon, for better accuracy is an important task for future work.

\section*{Acknowledgement}
The work was supported in part by National Key R\&D Program of China under Grant 2017YFE0121500, NSFC under Grant 61871441, 91738202, 61631015, Young Elite Scientists Sponsorship Program by CAST with Grant 2018QNRC001, Major Program of National Natural Science Foundation of Hubei in China with Grant 2016CFA009, Key Laboratory of Dynamic Cognitive System of Electromagnetic Spectrum Space (Nanjing Univ. Aeronaut. Astronaut.), MIIT, Fundamental Research Funds for the Central Universities with Grant number 2015ZDTD012, and Graduates' Innovation Fund of HUST with Grant 2019YGSCXCY009.

\bibliography{main}
\bibliographystyle{IEEEtran}

\end{document}